\documentclass[10pt,twocolumn,letterpaper]{article}

\usepackage[pagenumbers]{cvpr} 

\usepackage{graphicx}
\usepackage{amsmath}
\usepackage{xcolor}
\usepackage{colortbl}
\usepackage{amssymb}
\usepackage{booktabs}

\usepackage[pagebackref,breaklinks,colorlinks]{hyperref}

\usepackage[capitalize]{cleveref}
\crefname{section}{Sec.}{Secs.}
\Crefname{section}{Section}{Sections}
\Crefname{table}{Table}{Tables}
\crefname{table}{Tab.}{Tabs.}

\usepackage{bbding}
\usepackage{xcolor}
\definecolor{green}{RGB}{0,255,0}
\definecolor{red}{RGB}{255,0,0}
\usepackage{multirow}
\usepackage{stfloats}
\usepackage{bm}
\def\cvprPaperID{8383} 
\def\confName{CVPR}
\def\confYear{2023}

\begin{document}

\title{Window Normalization: Enhancing Point Cloud Understanding by Unifying Inconsistent Point Densities}

\author{
    Qi Wang$^{1,2}$,
    Sheng Shi$^{1}$\footnotemark[1],
    Jiahui Li$^{1}$,
    Wuming Jiang$^{3}$,
    Xiangde Zhang$^{1}$\\
    $^1$Northeastern University \\
    $^2$Key Laboratory of Data Analytics and Optimization for Smart Industry (Northeastern University) \\
    $^3$Beijing Eyecool Technology Co., Ltd. \\
}
\maketitle

\begin{abstract}
  Downsampling and feature extraction are essential procedures for 3D point cloud understanding.
  Existing methods are limited by the inconsistent point densities of different parts in the point cloud.
  In this work, we analyze the limitation of the downsampling stage and propose the pre-abstraction group-wise window-normalization module.
  In particular, the window-normalization method is leveraged to unify the point densities in different parts.
  Furthermore, the group-wise strategy is proposed to obtain multi-type features, including texture and spatial information.
  We also propose the pre-abstraction module to balance local and global features.
  Extensive experiments show that our module performs better on several tasks.
  In segmentation tasks on S3DIS (Area 5), the proposed module performs better on small object recognition, and the results have more precise boundaries than others.
  The recognition of the sofa and the column is improved from 69.2\% to 84.4\% and from 42.7\% to 48.7\%, respectively.
  The benchmarks are improved from 71.7\%/77.6\%/91.9\% (mIoU/mAcc/OA) to 72.2\%/78.2\%/91.4\%.
  The accuracies of 6-fold cross-validation on S3DIS are 77.6\%/85.8\%/91.7\%.
  It outperforms the best model PointNeXt-XL (74.9\%/83.0\%/90.3\%) by 2.7\% on mIoU and achieves state-of-the-art performance.
  The code and models are available at  \url{https://github.com/DBDXSS/Window-Normalization.git}.
\end{abstract}


\section{Introduction}
    \label{sec:1}
    3D point clouds have been widely researched. It has rich applications, such as autonomous driving, defect detection, and robotics. A point cloud is a set of coordinates of a real object in the 3D Cartesian coordinate system. The obtained representations include color, reflection intensity, and some others. Different from 2D images, point clouds are sparse and disordered. There are huge differences between different observations of the same object. These characteristics differ from 2D images, so point clouds and images are processed differently.

    In recent years, many networks have been proposed for processing point clouds. It can be divided into projection-based, voxel-based, and point-based methods. Projection-based methods \cite{1Su_2015,2Li_2016,3Chen_2017,4Kanezaki_2018,5Lang_2019,6Tatarchenko_2018} benefit from image-based methods. They transfer point clouds to 2D images via a projection-based model. Voxel-based methods \cite{7Maturana_2015,8Song_2017,9Riegler_2017,10Graham_2018,11Choy_2019} transform point clouds into regular dense structures and ues image methods  to further process the dense structures. However, projection-based and voxel-based methods inevitably suffer from the loss of spatial information. Point-based methods
    \cite{12Charles_2017,13Charles_2017,14Chu_2022,15Hu_2020,16Hu_2020,17Lei_2020,18Liu_2021,19Ye_2022,20Xu_2021,21Yan_2020,22Yang_2019,23Zhao_2019,24Zhao_2021,25Shao_2022} directly process the raw point cloud. Point-based methods have developed rapidly since PointNet++ \cite{13Charles_2017} proposed the farthest point sampling (FPS) method and hierarchical processing network.

\begin{figure*}[!htb]
    \centering
    \includegraphics[width=1\linewidth]{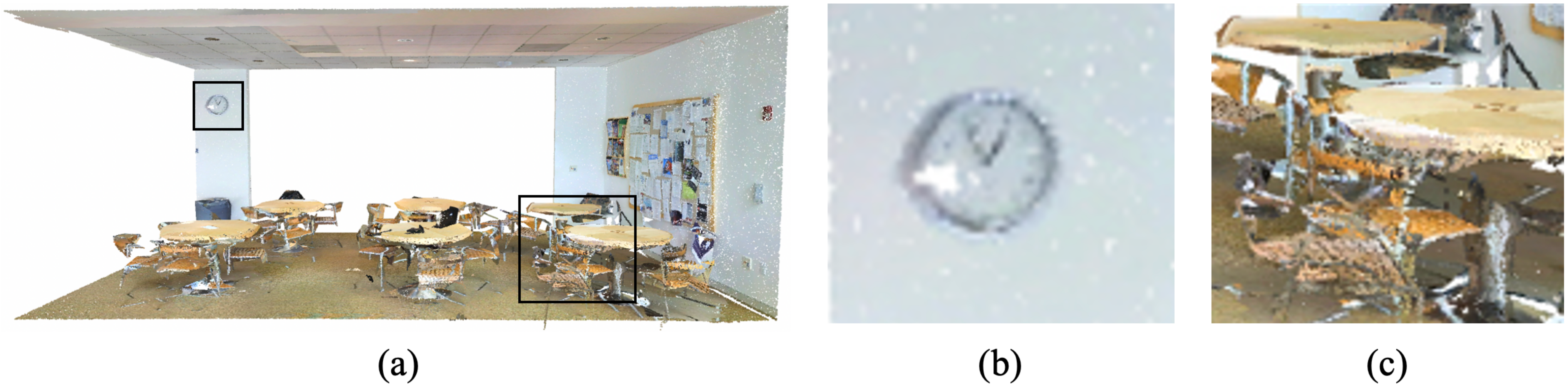}
    \caption{Different parts of the point cloud have different point densities. (a) Whole scene. (b) High-density part. (c) Low-density part.}
    \label{fig:1}
    \end{figure*}

    There are still some problems with point-based methods.
    As shown in Figure \ref{fig:1}, the different parts of the point cloud have different point densities.
    The local neighborhoods obtained by existing methods have different volumes.
    They share the same network weights, which is adverse to the convergence of the network.
    Moreover, only using the max-pooling function (MaxPool) can not balance local and global features well.
    We propose a pre-abstraction group-wise window-normalization (PAGWN) module for this situation to improve the performance of point-based semantic segmentation networks.
    A better balance between local and global features can be obtained.
    With the proposed method, it can improve sampling accuracy and reduce the information loss caused by the sampling stage, enabling point-based methods to perform better.

    Extensive experiments show that the PAGWN module performs better on several tasks.
    The proposed module performs better on small object recognition, and the results have more precise boundaries than others.
    In segmentation tasks on S3DIS (Area 5), the recognition of the sofa and the column is improved from 69.2\% to 84.4\% and from 42.7\% to 48.7\%, respectively.
    The benchmarks are improved from 71.7\%/77.6\%/91.9\% (mIoU/mAcc/OA) to 72.2\%/78.2\%/91.4\% and achieve state-of-the-art.
    The accuracies of 6-fold cross-validation on S3DIS are 77.6\%/85.8\%/91.7\% (mIoU/mAcc/OA). 
    It outperforms the best model PointNeXt-XL \cite{45https://doi.org/10.48550/arxiv.2206.04670} (74.9\%/83.0\%/90.3\% (mIoU/mAcc/OA)) by 2.7\% on mIoU and achieves state-of-the-art performance.
    Overall, our contributions are summarized below:
    \begin{itemize}
    \item We study the downsampling stage in a novel way. A general feature extractor PAGWN is proposed for point cloud understanding. It can better balance local and global information and reduce information loss.
    \item The proposed module performs better on small object recognition, and the segmentation results have more precise boundaries than others. Many existing networks can achieve better performances by using the pre-abstraction group-wise window-normalization module without making any other changes.
    \item The benchmarks of segmentation tasks on S3DIS (Area 5) are improved from 71.7\%/77.6\%/91.9\% (mIoU/mAcc/OA) to 72.2\%/78.2\%/91.4\%. The accuracies of 6-fold cross-validation on S3DIS are 77.6\%/85.8\%/91.7\%. It outperforms the best model PointNeXt-XL \cite{45https://doi.org/10.48550/arxiv.2206.04670} (74.9\%/ 83.0\%/90.3\%) by 2.7\% on mIoU and achieves state-of-the-art performance.
    \end{itemize}

\section{Related Work}

    The methods of processing point clouds can be divided into the following types: Convert the point clouds into 2D images by multi-view projection; Perform regular grid division on the point clouds and transform them to voxels or lattices; Process the point clouds directly based on points. The point-based approaches are based on the advantage of not doing too much processing on the point clouds. They try to keep the original features of the point clouds and avoid information losses caused by data processing. The point-based networks can be divided into point-based multi-layer perceptron (MLP) networks, point-based convolution networks, RNN-based networks, and graph-based networks. Among the point-based MLP networks, the methods used in local feature aggregation can be classified into pooling-based, attention-based, and transformer-based methods.

    \textbf{Pooling-based Methods}  The first point-based method is proposed by PointNet \cite{12Charles_2017}. It directly processes point clouds with symmetric functions such as MLP networks and pooling methods. PointNet++ \cite{13Charles_2017} proposes farthest point sampling. Moreover, a hierarchical extraction network is constructed based on PointNet. It gradually expands the receptive field of the network and improves the ability to aggregate local features. The proposal of FPS and the construction of hierarchical networks have brought new inspiration to point-based methods. Point-based methods have developed rapidly since PointNet++. PointSIFT \cite{26} encodes information in eight spatial directions through a three-stage ordered convolutional model. The multi-scale features are spliced to adapt to different scales. PointWeb \cite{23Zhao_2019} builds a fully connected graph by exploring the pairing relationships of points within a local region. PointMLP \cite{27} normalizes different samples, and local features are obtained by the concatenation method.

    \textbf{Self-attention-based Methods}  Attention mechanism \cite{28} is introduced to enhance the local feature extraction ability. Yang et al. \cite{22Yang_2019} propose a grouped random attention model to construct relationships between points. Gumbel Subset Sampling is proposed to replace FPS. This sampling method is insensitive to contours and can select more representative points. Local Spatial Aware layer \cite{29} and Attention-based Score Refinement module \cite{30Zhao_2019} are proposed to learn spatial weights of local structure in point clouds. RandLA-Net \cite{15Hu_2020} proposes random sampling for fast sampling and improves sampling efficiency. It uses Cartesian coordinates and points' features for stitching to learn spatial weights, and pooling is used for local feature aggregation.

    \textbf{Transformer-based Methods}  Following the success of the Transformer structure in the vision domain \cite{31Thomas_2019,32Carion_2020,33nbcaaa,34qq,35qq,36Liu_2021,37Mao_2021,38qq,Touvron,40Touvron_2021,41Wang_2022}, many studies \cite{24Zhao_2021,42Engel_2021,43Guo_2021,44qq} have begun to use it for processing point clouds. Engel et al. \cite{42Engel_2021} utilize SortNet and Global Feature Generation to extract local and global features. The global and local features are integrated using local-global attention. Guo et al. \cite{43Guo_2021} use Transformer for feature extraction based on PointNet++. Offset-Attention is proposed to form a residual structure to improve the accuracy further. Zhao et al. \cite{24Zhao_2021} propose using positional encoding twice to improve the Transformer effect. Lai et al. \cite{44qq} propose a StratifiedFormer to obtain long-range contextual information and achieve better performance.

    To reduce the information loss, we propose the PAGWN module to unify the point densities.
    With the PAGWN module, the original network can perform better and maintain its structure unchanged.

\section{Methods}
    \subsection{Overview}
\begin{figure*}[!htb]
        \centering
        \includegraphics[width=0.8\linewidth]{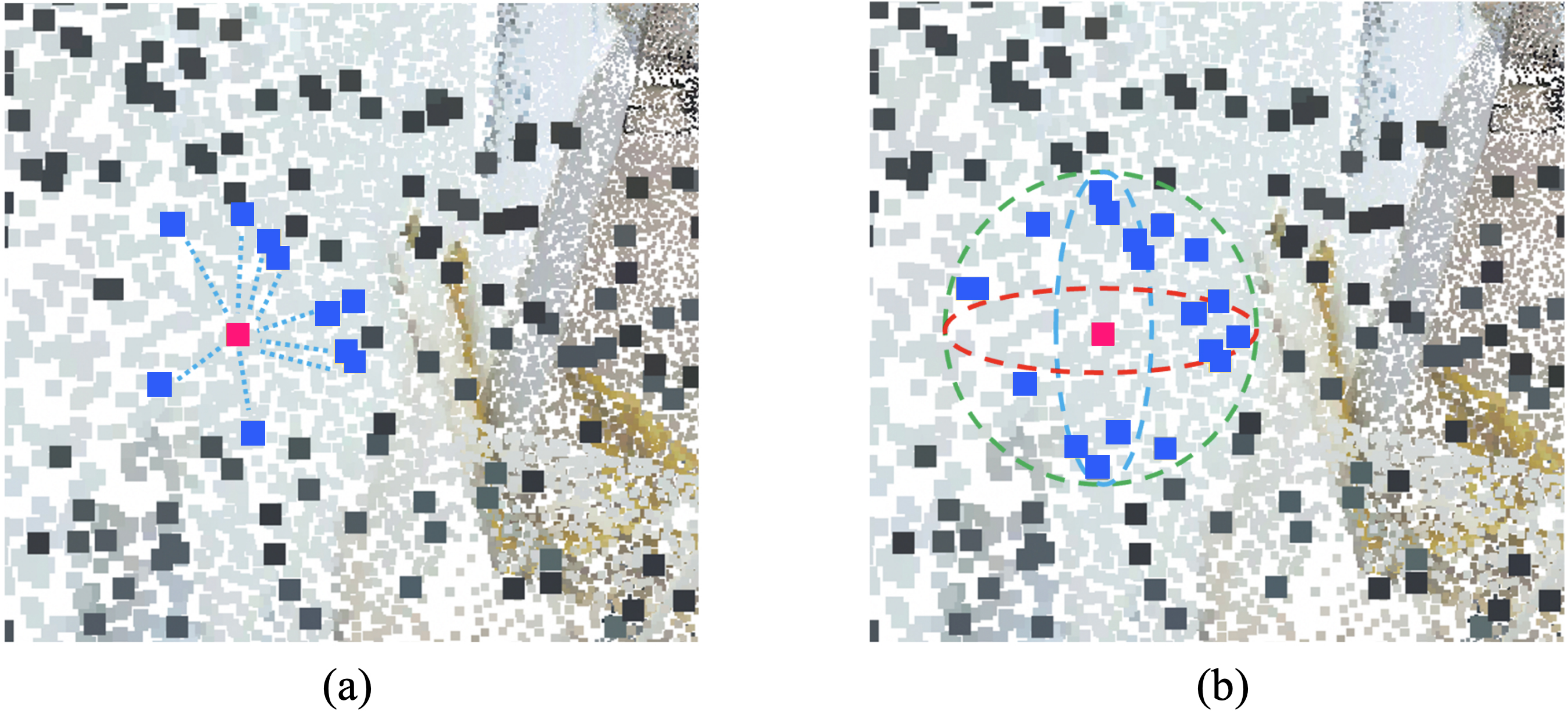}
        \caption{Different methods to obtain neighbor points. The neighborhood obtained by the KNN method has a fixed number of points but cannot limit the neighborhood volume. The neighborhood obtained by the BQ method has a fixed volume but no fixed number. (a) uses the $k$-nearest neighbor method to gain local region, while (b) uses ball query.}
        \label{fig:2}
        \end{figure*}

        Downsampling and feature extraction are essential procedures for 3D point cloud understanding. Existing methods are limited by the inconsistent point densities of different parts in the point cloud. The downsampling stage aims to enlarge the receptive field by reducing the point cloud size without losing too much information. There are two steps in the downsampling stage. In the first step, select a certain number of points to represent the entire point cloud, which commonly uses random sampling, FPS, and other methods. The second step is performing local feature aggregation on the sampled point cloud. It aims to enhance the information of the point cloud. Generally, the $k$-nearest neighbor (KNN) method or ball query (BQ) \cite{13Charles_2017} is used to select neighbor points. Then MLP and MaxPool are used for local feature aggregation. Figure \ref{fig:2} illustrates the different methods: KNN and BQ.

\begin{figure*}[!htb]
        \centering
        \includegraphics[width=0.8\linewidth]{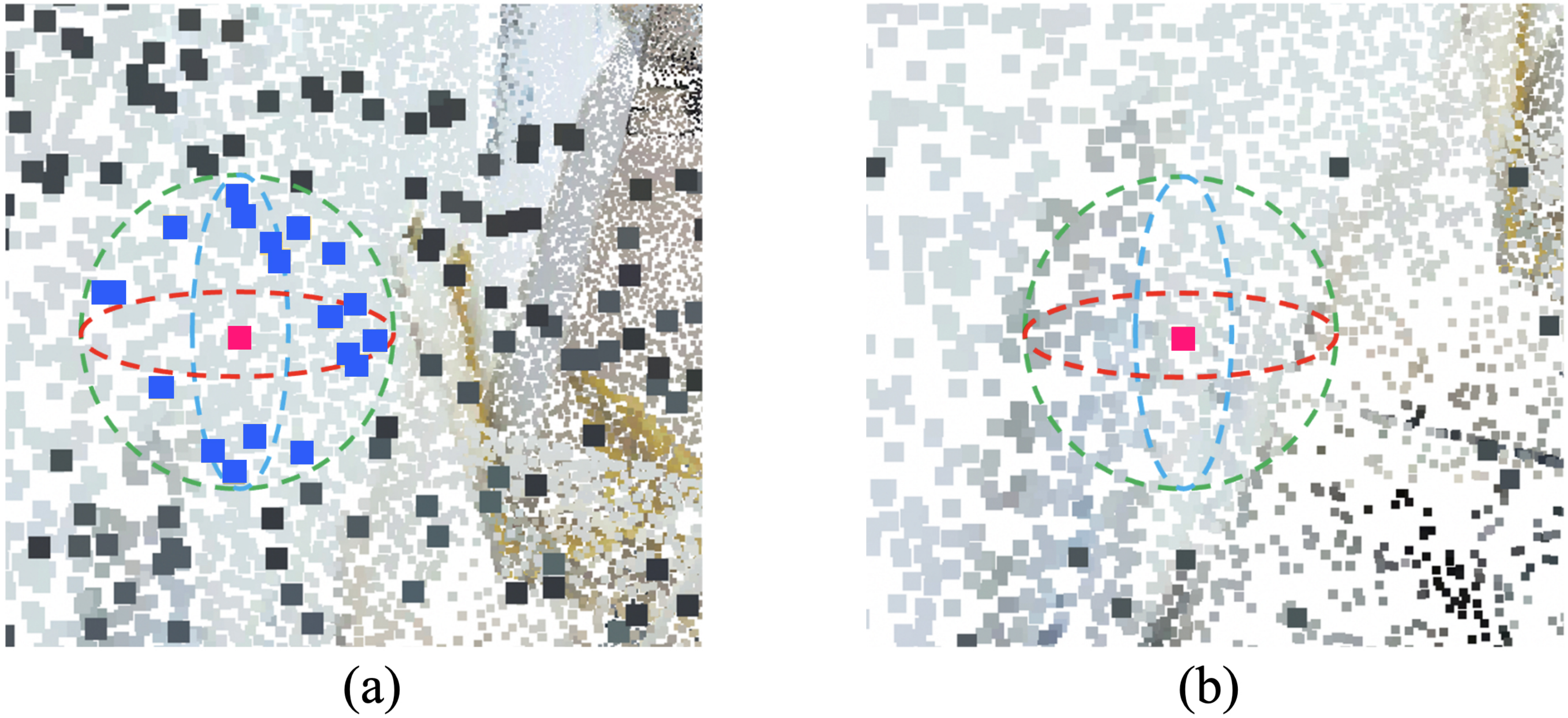}
        \caption{Different parts with different point densities have a different number of neighbor points with BQ on a fixed scale. (a) The BQ method will produce more neighbor points in high-density parts. (b) If the density is too low, there will be fewer, even no, points in the local region.}
        \label{fig:3}
        \end{figure*}

        With the development of hierarchical networks, the downsampling stage has been unavoidable in many existing methods to enlarge the receptive field effectively. Suppose $P$ is the given point cloud. $P = \{ ({c_i},{x_i}) \vert i = 1,\cdots,N\}  \in {R^{N \times (3 + n)}}$. Here ${c_i}$ is the 3D Cartesian coordinate of the $i$-th point. ${x_i}$ is the $n$-dimensional point cloud feature of the $i$-th point. $N$ is the number of points contained in the point cloud. In PointNet++ \cite{13Charles_2017}, FPS is used for point cloud downsampling. BQ is used for local point cloud extraction. MLP and MaxPool are used for local feature aggregation:
        \begin{equation}
        \begin{aligned}
        {\hat x_{i}} = &MaxPool\{ MLP\left( {[{x_{i,j}}]} \right)\} ; \\ \forall {x_i} \in FPS\left( P \right)&; {x_{i,j}} \in BQ\left( {{x_i}} \right),j = 1, \cdots ,K,
        \end{aligned}
        \end{equation}
        where MLP comprises a fully-connected layer, a batch normalization layer, and an activation function. There are $K$ neighbor points in total. ${x_i}$ is the $i$-th point in the sampled point cloud. ${x_{i,j}}$ is the $j$-th neighbor of the $i$-th point. PointNet++ \cite{13Charles_2017} proposes that the local region obtained by BQ with a fixed scale is beneficial to learn local patterns. However, as shown in Figure \ref{fig:3}, BQ cannot solve the problem of different point densities in different parts of the point cloud. If there are too many points in the local region, the amount of computation will be enormous. Furthermore, if there are too few points, the estimate of the center point will be easily biased. So multi-scale grouping (MSG) and multi-resolution grouping (MSR) are further proposed to extract multi-scale features and improve accuracy.

        With the increment in model complexity and data scale, MSG and MSR can no longer meet the requirements for computational efficiency. However, using single-scale grouping will reduce the sampling accuracy. So, Point Transformer \cite{24Zhao_2021} and StratifiedFormer \cite{44qq} use KNN to replace BQ:
        \begin{equation}
        \begin{aligned}
        {\hat x_{i}} = &MaxPool\{ MLP\left( {[{x_{i,j}}]} \right)\} ; \\ \forall {x_i} \in FPS\left( P \right)&;{x_{i,j}} \in KNN\left( {{x_i}} \right),j = 1, \cdots ,K.
        \end{aligned}
        \end{equation}

        However, the local region sampled by KNN has no fixed scale. Different parts with different volumes share the same network weights, which is adverse to the convergence of the network. Therefore, it is necessary to propose a more efficient sampling method based on KNN, which motivates us to conduct further studies to improve the accuracy of the downsampling stage. We wish to develop a general downsampling scheme that works for segmentation networks.

    \subsection{Window-normalization}
        Let us consider the problem of different scales of local regions obtained by KNN. Existing normalization methods include Layernorm and Batchnorm. The Batchnorm balances the scale between different samples, and the Layernorm normalizes the features of a single instance. They both can not solve the problem. So, we propose a new normalization method--window normalization with prior knowledge.

\begin{figure*}[!htb]
        \centering
        \includegraphics[width=0.65\linewidth]{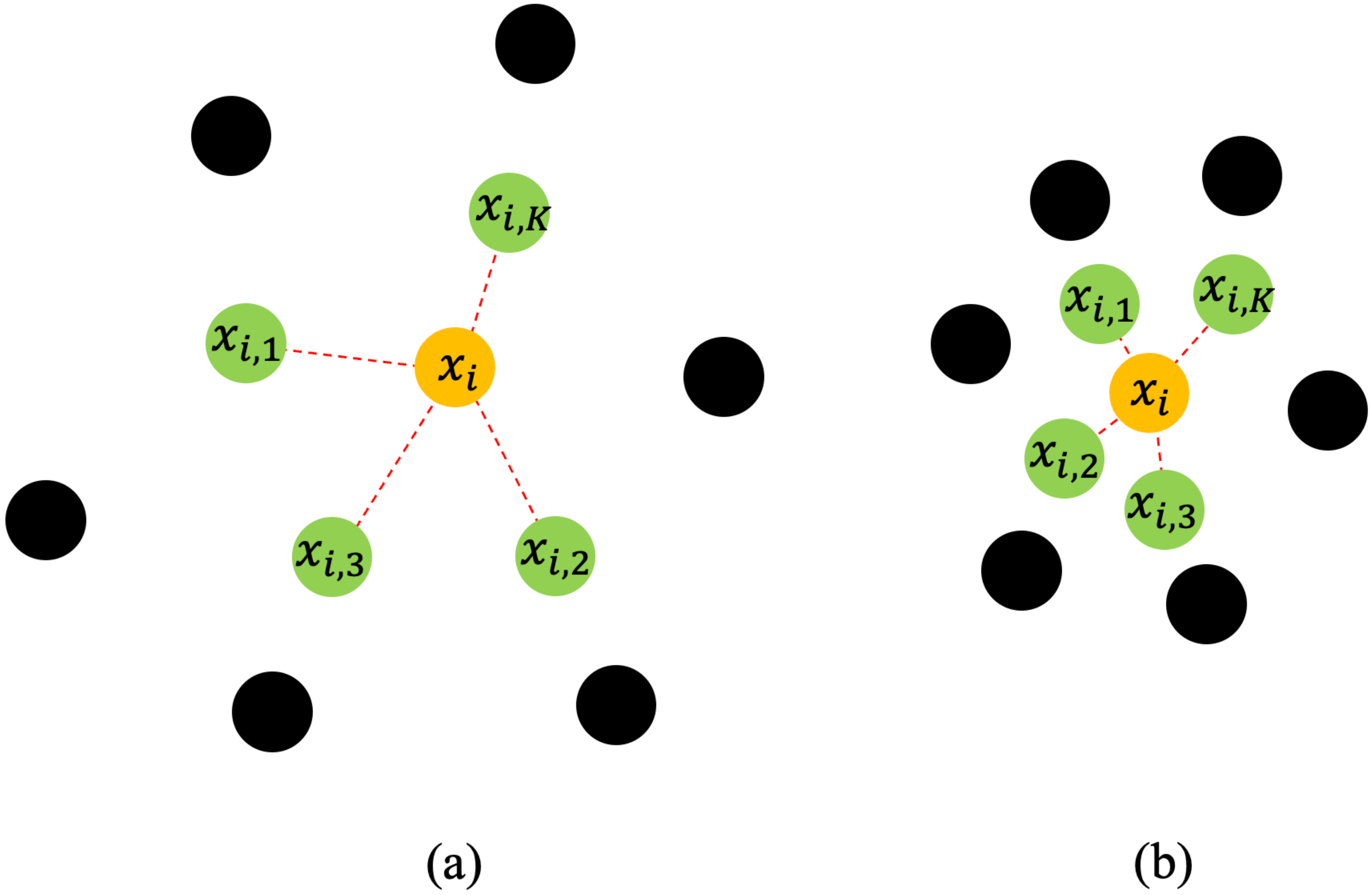}
        \caption{Apply KNN on point clouds with different point densities. Neighbors obtained in low-density regions have a larger volume but the volume is smaller in high-density regions. (a) Low-density region. (b) High-density region.}
        \label{fig:4}
        \end{figure*}

        For a point cloud $P$ with $N$ points, a point cloud $Q$ ($Q \subset P$) with $M$ ($M<N$) points is firstly obtained by the sampling stage. Then, the $K$ nearest points to $x \in Q$ are selected from the original point cloud $P$ as the local region to improve the sampling accuracy. The feature aggregation is performed on the local region to obtain a new sampled point ${\hat x} $. As shown in Figure \ref{fig:4}, the local region with $K$ points is obtained by the KNN method. Due to the different densities of different parts in the point cloud, the volumes occupied by the $K$ points are quite different. Most existing aggregation methods use the MLP network to learn local features directly and use the max-pooling function for feature aggregation. However, different local regions sharing the same weight will produce aggregation loss.

        Neither Batchnorm nor Layernorm can solve this problem. Batchnorm unifies the scales of different samples, which cannot solve this problem. Layernorm normalizes individual samples. Local regions need to be normalized as a single sample to solve regional scale inconsistency.

\begin{figure}[!htb]
        \centering
        \includegraphics[width=0.5\linewidth]{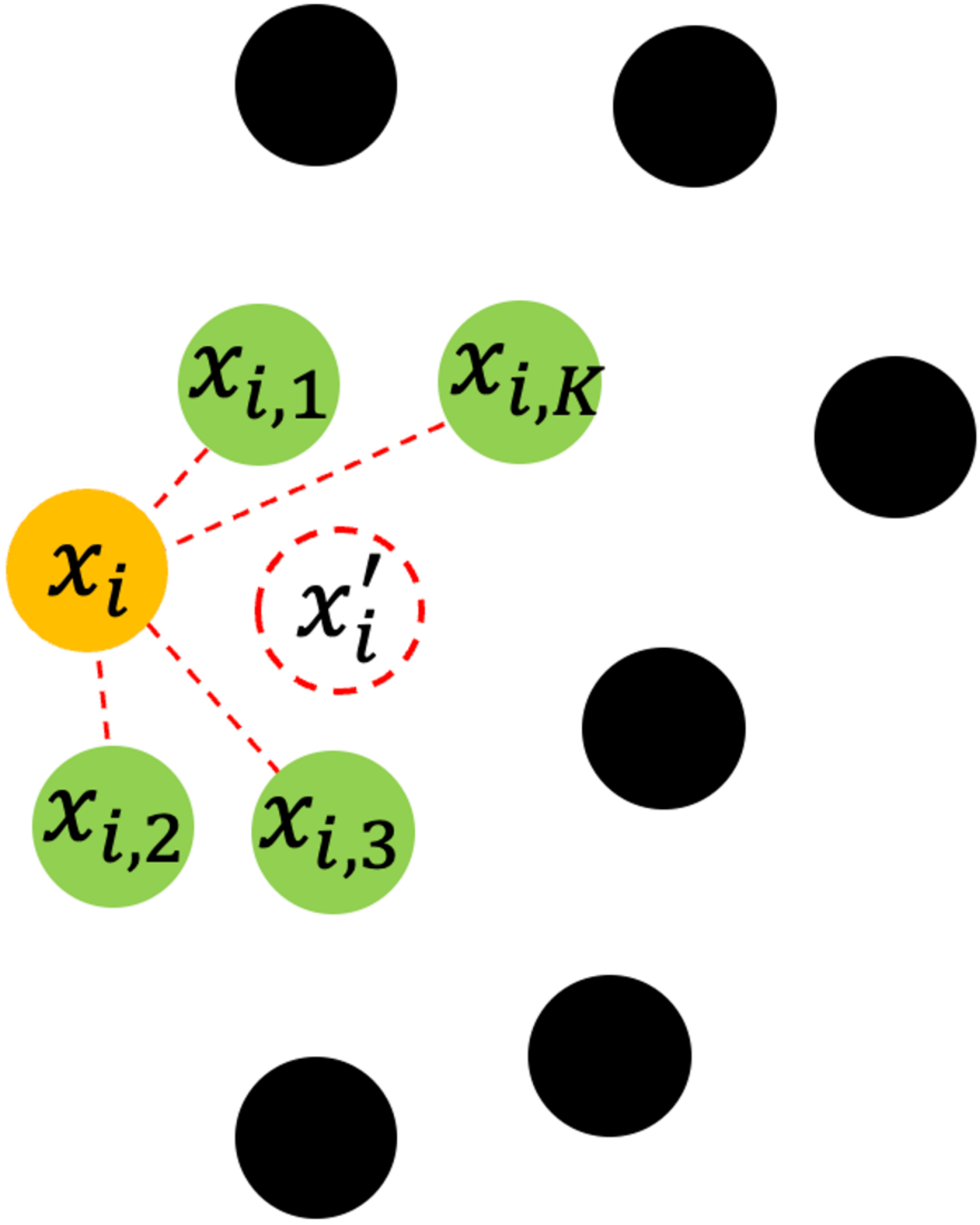}
        \caption{The expectation of the local region by KNN is different from the center point.}
        \label{fig:5}
        \end{figure}

        In addition to the errors caused by different neighborhood volumes, the normalization method also needs further improvement. Affected by KNN, the center of the sampled region is probably not the sampling point. Expectations will be biased. As shown in Figure \ref{fig:5}, the $K$ neighbor points sampled from the point cloud are all on one side of the center point. They are not evenly distributed around the center point. The features obtained by center normalization will deviate from the center point.

\begin{figure*}[!htb]
        \centering
        \includegraphics[width=0.5\linewidth]{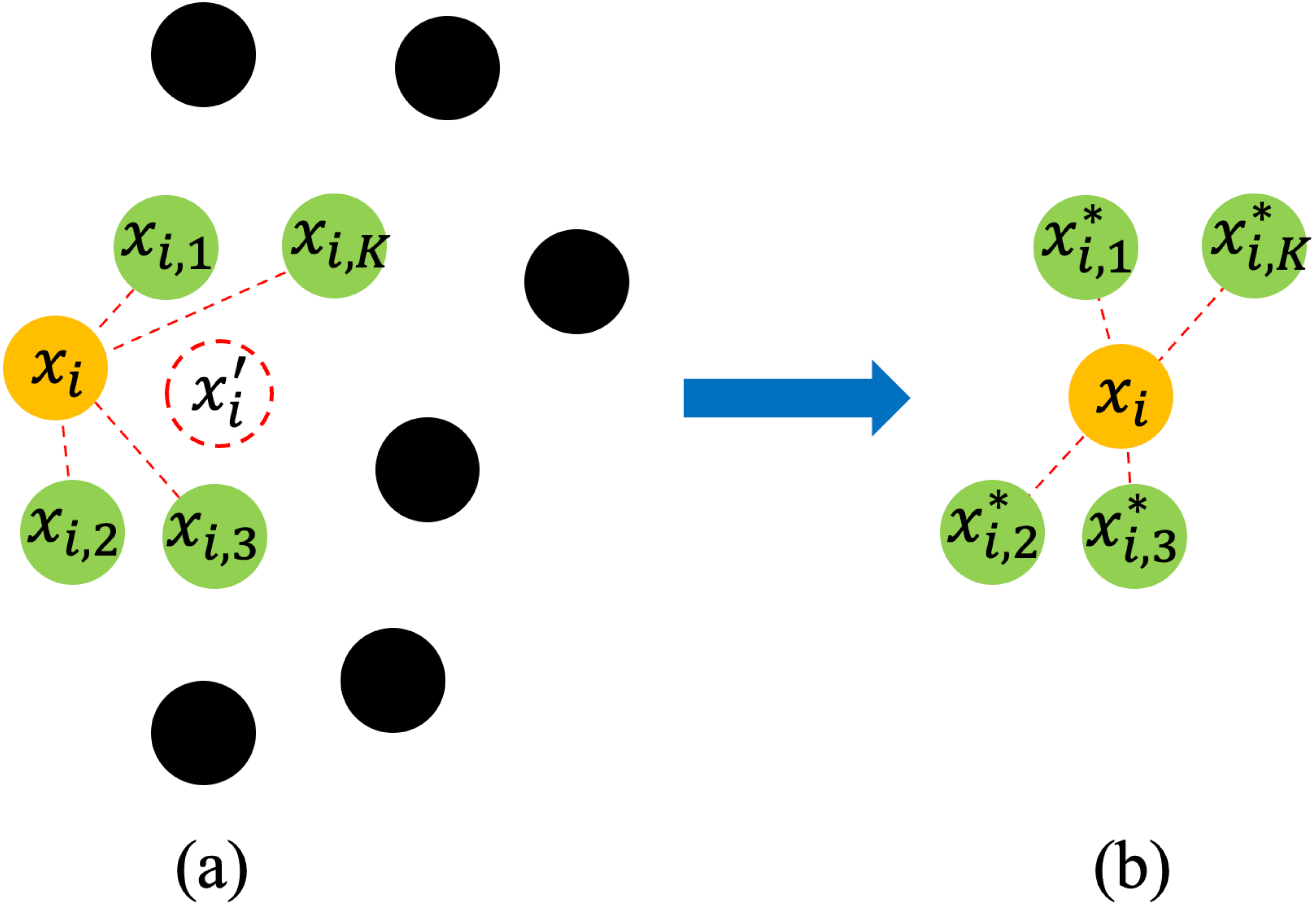}
        \caption{We use the window-normalization method with prior knowledge to calibrate local features. (a) Original expectation by KNN. (b) Calibrated expectation by proposed method}
        \label{fig:6}
        \end{figure*}

        Inspired by Layernorm and Batchnorm, a window normalization method that uses prior knowledge is proposed. It is used to normalize the features of the local region. It can accelerate the learning process and improve the network's accuracy. We normalize the neighborhoods obtained by different sampling regions and achieve the unification of neighborhood features. Based on prior knowledge, the center point is taken as the mean point instead of using the mean value of point features in the neighborhood. It can maintain global features and reduce the offset problem caused by sampling. As shown in Figure \ref{fig:6}, ${x_i}$ is the center point. ${x'_i}$ is the mean of the sampling points. ${x_{i,k}}$ is the $k$-th neighbor point of the $i$-th center point. Window-normalization (windownorm) follows the normalization method, where the features are subtracted from the expectation and divided by the standard deviation. The expectation within the local region should be the feature of the sampling point. The variance used by the normalization is the standard deviation of all features in the local region. The normalization formula can be expressed as:
        \begin{equation}
          \begin{split}
        &{\hat x_{i,j}} = \frac{{{x_{i,j}} - {x_i}}}{{\sigma  + \varepsilon }},\;\\
        \sigma = &\sqrt {\frac{1}{{K \times n - 1}}\mathop \sum \limits_{j = 1}^K {{\left( {{x_{i,j}} - {x_i}} \right)}^2}} ;\; \\
        \forall {x_i} \in FPS&\left( P \right);{x_{i,j}} \in KNN\left( {{x_i}} \right), j=1, \cdots ,K,
          \end{split}
        \end{equation}
        where $\sigma $ is the standard deviation. ${\varepsilon  = 1e - 5}$ is a small constant used to guarantee the stability of the calculation. $n$ is the feature dimension. ${{x_i}}$ is the center point. ${{x_{i,j}}}$ is the $j$-th neighbor of ${{x_i}}$. Window normalization solves the problem of different window scales without any learnable parameters. The window itself determines the different window expectations and variances. The calibration formulae can be described as follows:
        \begin{equation}
            \begin{aligned}
        x_{i,j}^* &= \hat x_{i,j} + {x_i} = {{{x_{i,j}} - {x_i}} \over {\sigma  + \varepsilon }} + {x_i}  \\
        &= {{{x_{i,j}}} \over {\sigma  + \varepsilon }} + \left( {1 - {1 \over {\sigma  + \varepsilon }}} \right) \cdot {x_i} \\
        &= \lambda \cdot {x_{i,j}} + \left( 1 - \lambda \right) \cdot {x_{i}}, \ \lambda = {1 \over {\sigma + \varepsilon}};
            \end{aligned}
        \end{equation}
        \begin{equation}
            \begin{aligned}
        E\left( {x_{i,j}^*} \right) &= {{E\left( {{x_{i,j}}} \right) - {x_i}} \over {\sigma  + \varepsilon }} + {x_i} \\
        &= {\lambda} \cdot E\left( {{x_{i,j}}} \right) + \left( 1-{\lambda} \right) \cdot {x_{i}};
            \end{aligned}
        \end{equation}
        \begin{equation}
            \begin{aligned}
        Var\left( {x_{i,j}^*} \right) &= {\left( {{1 \over {\sigma  + \varepsilon }}} \right)^2} \cdot Var\left( {{x_{i,j}}} \right) \\
        &= {\lambda}^2 \cdot Var\left( {{x_{i,j}}} \right).
            \end{aligned}
        \end{equation}
        where ${x_{i,j}^*}$ is the rectified point of ${x_{i,j}}$.

        It can be seen that if the original feature is unbiased, the corrected feature will still be unbiased.
        If the standard deviation is larger, the local point cloud would be more likely to be sparse.
        Moreover, the rectified neighbor point features will be more unbiased and practical.
        Furthermore, if there are few points around the center point, less distinguishable information can be provided in the neighborhood of the center point, leading to difficulties in classification.
        The proposed method can improve the unbiasedness and effectiveness of such regions, which is beneficial for segmenting complex objects.

\begin{figure*}[!htb]
        \centering
        \includegraphics[width=1\linewidth]{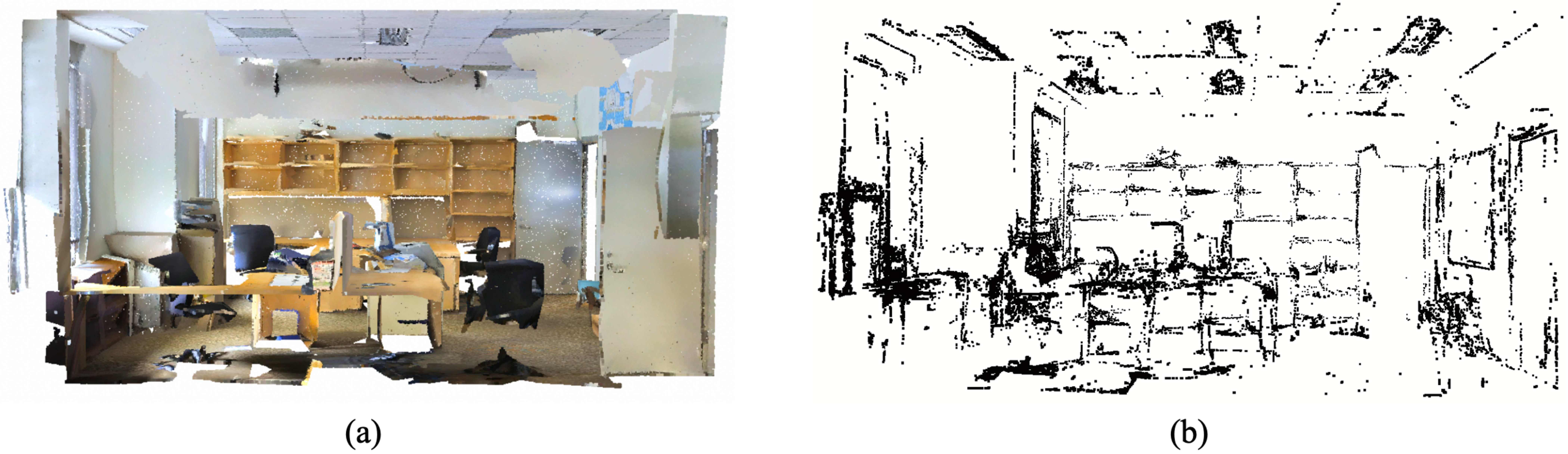}
        \caption{The standard deviation of different regions in the point cloud. The center points with the standard deviation of neighbors larger than 1 are displayed. (a) A point cloud with coordinates and RGB features of a scene in S3DIS (Area 5). (b) Points with the standard deviation of neighbors larger than 1.}
        \label{fig:7}
        \end{figure*}

        To show the sparsity in natural scenes, we calculate the standard deviation of different regions in the point cloud.
        The center points with the standard deviation larger than 1 are displayed in Figure \ref{fig:7}.
        As can be seen, the extracted points are almost boundaries, which is difficult for semantic segmentation.

    \subsection{Group-wise Window-normalization}

\begin{figure}[!htb]
        \centering
        \includegraphics[width=0.8\linewidth]{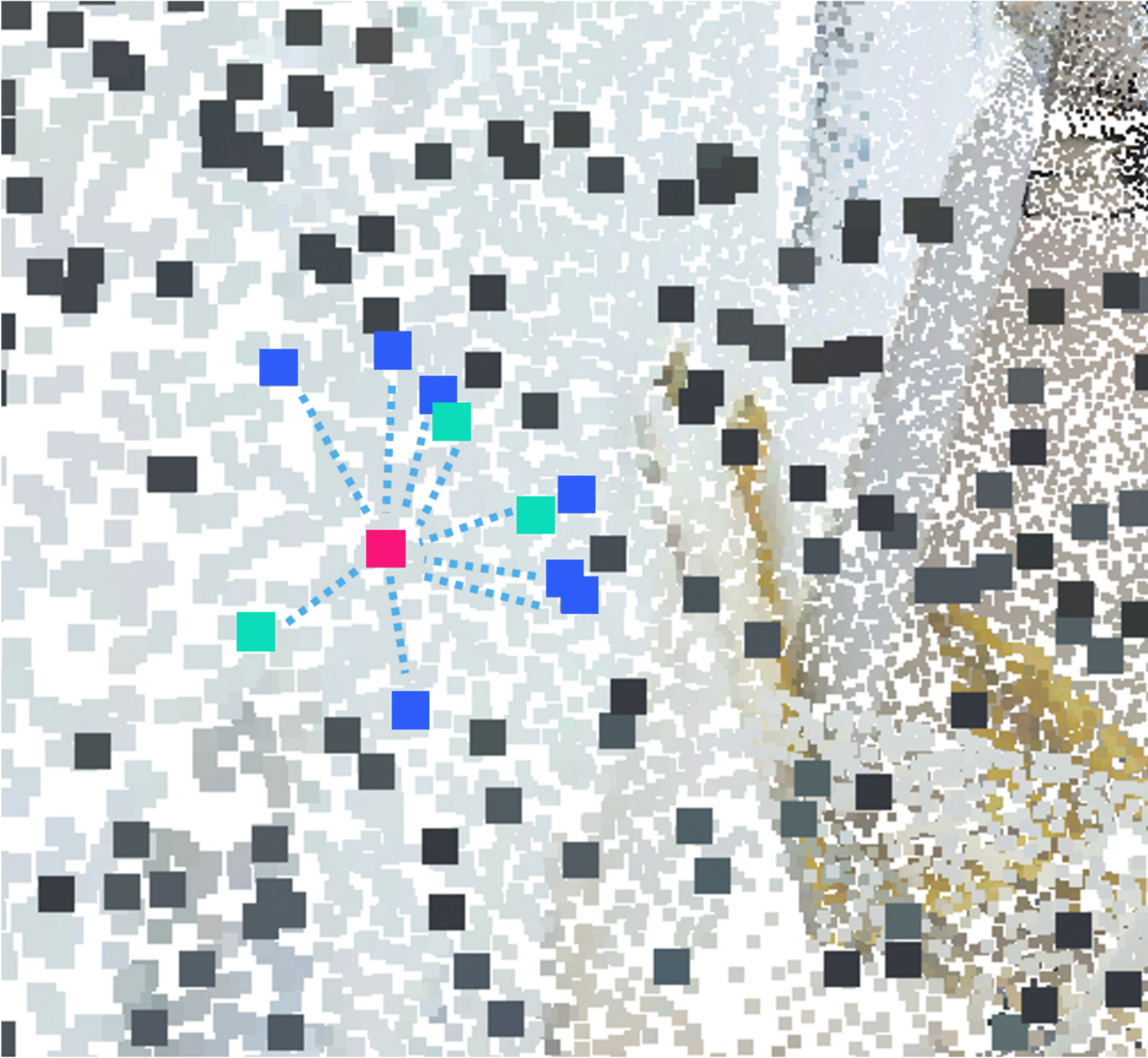}
        \caption{In the local region, green points are closer to the center point, providing texture information that enhances the local features. Blue points are far from the center point, providing spatial information with a bigger receptive field.}
        \label{fig:8}
        \end{figure}

        As shown in Figure \ref{fig:8}, we argue that points in the neighborhood express two different kinds of feature information: texture information that enhances the local features of the center point and spatial information that increases the receptive field. Therefore, we divide the neighbor points into two groups for windownorm. The $m$ neighbors closest to the center point are used as the first set of points. Besides, the remaining neighbor points are used as the second set of points. Windownorm is used for the two sets of points respectively.

        With the group-wise Window-normalization module, the scales of two sets of points are unified.
        The first set of points enhances the texture information.
        And the second set provides a larger receptive field and richer spatial information.
        The formula is:
        \begin{equation}
            \begin{aligned}
        &{\hat x_{i,j}} = \frac{{{x_{i,j}} - {x_i}}}{{\sigma  + \varepsilon }};\forall {x_i} \in FPS\left( P \right);
        \\&{x_{i,j}} \in KNN\left( {{x_i}} \right),j = 1, \cdots ,K;\\
        \sigma = &\left\{ {\begin{array}{*{20}{c}}
        {\sqrt {\frac{1}{{m \times n - 1}}\mathop \sum \limits_{j = 1}^m {{\left( {{x_{i,j}} - {x_i}} \right)}^2}};}\\
        {\sqrt {\frac{1}{{\left( {K - m} \right) \times n - 1}}\mathop \sum \limits_{j = m + 1}^K {{\left( {{x_{i,j}} - {x_i}} \right)}^2}}.}
        \end{array}} \right.
            \end{aligned}
        \end{equation}

    \subsection{Pre-Abstraction Group-wise Window-normalization Module}

        When solving semantic segmentation problems, we usually supplement the spatial information of points with position encoding.
        However, position encoding and sampling points are both used as spatial information to enhance the features of the center point.
        The position encoding makes the feature dimension of neighbors larger than that of the center point.
        It leads the model learning to be more biasedness toward position information.
        We hope the sampled features include local and global information in the sampling stage.

        To balance local and global features, we propose a pre-processing module. It can pre-aggregate the features of location encoding and neighborhood points. The model can be expressed as:
        \begin{equation}
            \begin{aligned}
        {\hat x_{i,j}}& = LB\left( {GWN\left( {\left[ {{c_{i,j}},{x_{i,j}}} \right]} \right)} \right); \forall {x_i} \in FPS\left( P \right); \\
        &\left( {{c_{i,j}},{x_{i,j}}} \right) \in KNN\left( {{x_i}} \right),j = 1, \cdots ,K,
            \end{aligned}
        \end{equation}
        where GWN is the group-wise window-normalization module,
        $\left[  \cdot  \right]$ is the concatenation operation. $LB$ consists of a linear layer and a Batchnorm layer. The input and output feature dimension of the linear layer is $n$+3 and $n$, respectively. The pre-processing module efficiently reduces the spatial feature dimension to $n$. So, the neighborhood feature dimension is the same as the center point.

\begin{figure*}[!htb]
        \centering
        \includegraphics[width=1\linewidth]{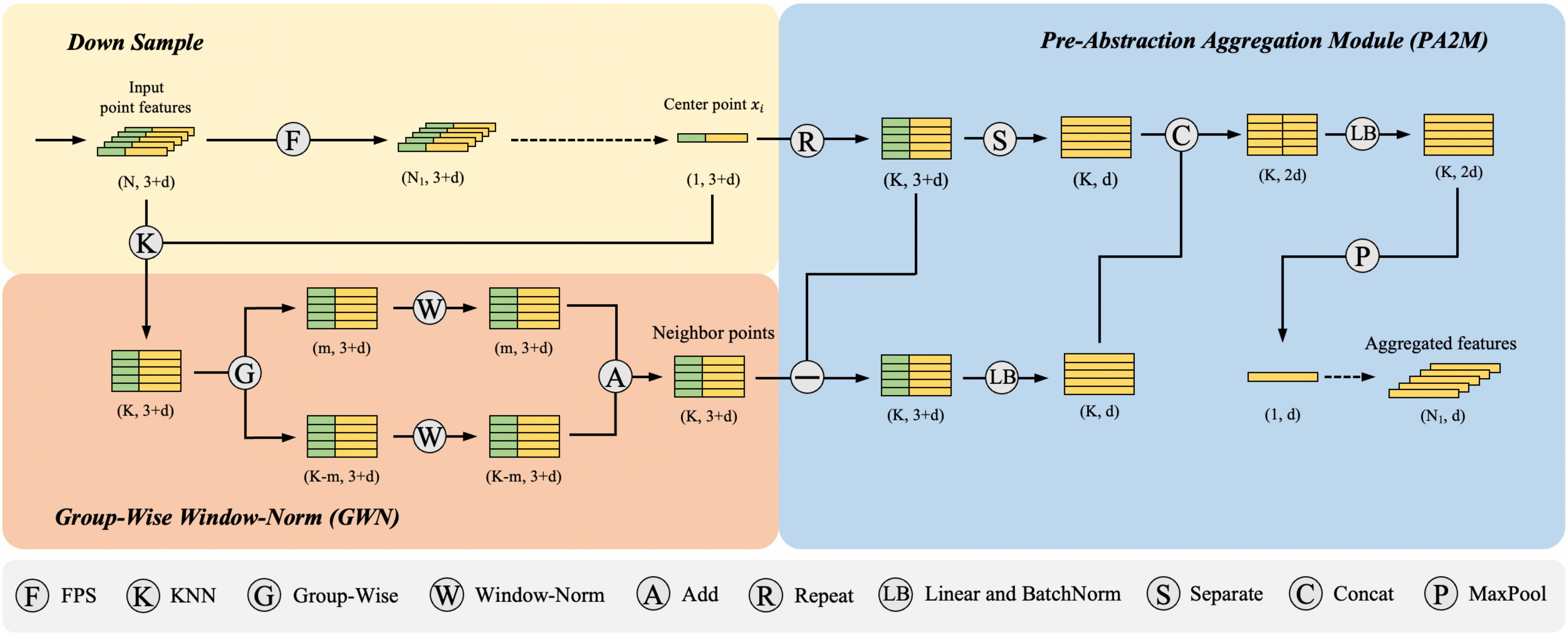}
        \caption{Illustration of our pre-abstraction group-wise window-normalization module. GWN module normalizes the volume of the local region and provides a variety of information. PAGWN module pre-aggregates the features of location encoding and neighborhood points to balance local and global features.}
        \label{fig:9}
        \end{figure*}

        The structure of the PAGWN module is shown in Figure \ref{fig:9}.
        The following formula can express the overall structure of the proposed sampling method:
        \begin{equation}
            \begin{aligned}
        x_{i,j}^{*} = {L{B_2}\left( {\left[ {L{B_1}\left( {GWN\left( {\left[ {{c_{i,j}},{x_{i,j}}} \right]} \right)} \right),{x_i}} \right]} \right)}\\
        {\hat x_{i}} = ReLU\left( {\mathop {{\rm{Maxpool}}}\limits_{j = 1, \cdots ,K} \left\{ x_{i,j}^{*} \right\}} \right); \forall {x_i} \in FPS\left( P \right); \\
        {\left( {{c_{i,j}},{x_{i,j}}} \right)} \in KNN\left( {{x_i}} \right),j = 1, \cdots ,K,
            \end{aligned}
        \end{equation}
        where the structure of ${L{B_1}}$ and ${L{B_2}}$ is the same as that of $LB$.
        The input feature dimension of ${L{B_1}}$ is $n$+3, and the output feature dimension is $n$.
        ${L{B_1}}$ is the pre-processing of local spatial information.
        The input and output feature dimensions of ${L{B_2}}$ are 2$n$.
        ${L{B_2}}$ is the aggregation of center and neighbor point information, which is used to aggregate local and global information.

\section{Experiments}
    \subsection{Semantic Segmentation on S3DIS (Area 5)}
        We evaluate the effectiveness of the proposed PAGWN module on several networks and tasks. The Point Transformer \cite{24Zhao_2021} and StratifiedFormer \cite{44qq} are used as the backbone, respectively.

        \textbf{Datasets}  The S3DIS dataset contains 271 rooms from three different buildings in six regions with 13 categories \cite{48Armeni_2016}. We test our networks on Area 5 and train on other areas. One RTX 3080Ti GPU is used to train networks. Mean intersection over union (mIoU), mean of class-wise accuracy (mAcc), and overall pointwise accuracy (OA) are used as evaluation metrics.

        \textbf{Setup}  We use Point Transformer and StratifiedFormer as the backbone, respectively. All settings of the dataset are the same as the backbone. We replace its downsampling module with our PAGWN module. The rest parameters are set according to the repositories \cite{46,47}.

\begin{figure*}[!htb]
            \centering
            \includegraphics[width=0.8\linewidth]{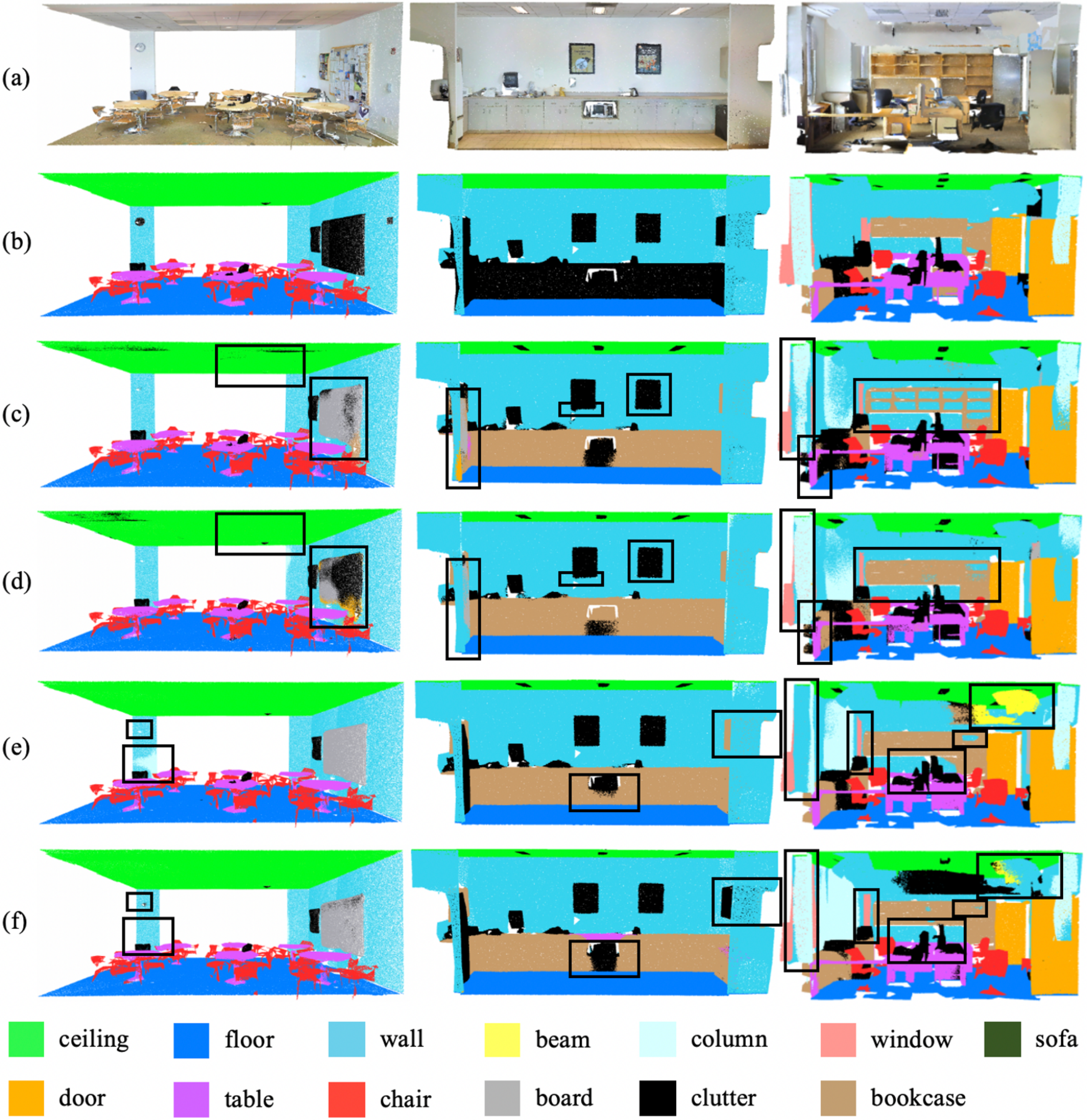}
            \caption{Quality comparison on three different scenes of S3DIS (Area 5). There are thirteen different kinds of categories colorized by different colors. We use black boxes to highlight where our model is better than the original. The proposed module performs better on small object recognition, and the results have more precise boundaries than others. Only our proposed module can locate the clock, cabinet, and bookcase in all tested networks. (a) Point clouds with coordinates and RGB features. (b) Ground truth Semantic labels. (c) Point Transformer. (d) Point Transformer with PAGWN. (e) StratifiedFormer. (f) StratifiedFormer with PAGWN.}
            \label{fig:10}
            \end{figure*}

        \textbf{Results}  We re-experiment Point Transformer and StratifiedFormer to ensure the results are compared under the same conditions.
        After that, we experiment with the Point Transformer first.
        The accuracies are improved from 70.6\%/77.1\%/90.5\% (mIoU/mAcc/OA) to 71.4\%/77.9\%/91.1\%.
        It surpasses PointNeXt \cite{45https://doi.org/10.48550/arxiv.2206.04670} to reach second.
        Furthermore, based on the StratifiedFormer, the recognition of the sofa and the column are improved from 69.2\% to 84.4\% and from 42.7\% to 48.7\%, respectively.
        The accuracies on StratifiedFormer are improved from 71.7\%/77.6\%/91.9\% (mIoU/mAcc/OA) to 72.2\%/78.2\%/91.4\% and achieve state-of-the-art.
        Figure \ref{fig:10} compares the quality of the semantic segmentation results of Point Transformer, StratifiedFormer, and our model.
        The specific results are shown in Table \ref{tab:1}.
        Extensive experiments show that networks can achieve better performance with the PAGWN module without modifying the other network structure, parameters, and training strategy.
        The proposed module performs better on small object recognition, and the results have more precise boundaries than others.

        \begin{table*}[!htb]\scriptsize\centering
            \renewcommand\arraystretch{1.3}
            \setlength{\tabcolsep}{1.5mm}
            \caption{Results of different models on S3DIS (Area 5, \%).}
            \begin{tabular}{c|ccc|ccccccccccccc}
            \toprule
            Method                                                                                                                      & mIoU & mAcc & OA     & ceiling & floor & wall & beam & column & window & door & table & chair & sofa & bookcase & board & clutter \\ \hline
            \begin{tabular}[c]{@{}c@{}}PointNet \cite{12Charles_2017}\end{tabular}                                                  & 41.1 & 49.0 & -      & 88.8    & 97.3  & 69.8 & 0.1  & 3.9    & 46.3   & 10.8 & 59.0  & 52.6  & 5.9  & 40.3     & 26.4  & 33.2    \\
            \begin{tabular}[c]{@{}c@{}}PAT \cite{22Yang_2019}\end{tabular}                                                          & 60.1 & 70.8 & -      & 93.0    & 98.5  & 72.3 & \textbf{1.0}  & 41.5   & \textbf{85.1}   & 38.2 & 57.7  & 83.6  & 48.1 & 67.0     & 61.3  & 33.6    \\
            \begin{tabular}[c]{@{}c@{}}PointWeb \cite{23Zhao_2019}\end{tabular}                                                     & 60.3 & 66.6 & 87.0   & 92.0    & 98.5  & 79.4 & 0.0  & 21.1   & 59.7   & 34.8 & 76.3  & 88.3  & 46.9 & 69.3     & 64.9  & 52.5    \\
            \begin{tabular}[c]{@{}c@{}}KPConv  \cite{31Thomas_2019}\end{tabular}                                                     & 67.1 & 72.8 & -      & 92.8    & 97.3  & 82.4 & 0.0  & 23.9   & 58.0   & 69.0 & 81.5  & 91.0  & 75.4 & 75.3     & 66.7  & 58.9    \\
            \begin{tabular}[c]{@{}c@{}}Point Transformer  \cite{24Zhao_2021}\end{tabular}                                           & 70.4 & 76.5 & 90.8   & 94.0    & 98.5  & 86.3 & 0.0  & 38.0   & 63.4   & 74.3 & \textbf{89.1}  & 82.4  & 74.3 & \textbf{80.2}     & 76.0  & 59.3    \\
            \begin{tabular}[c]{@{}c@{}}PointNeXt \cite{45https://doi.org/10.48550/arxiv.2206.04670}\end{tabular}                  & 70.5 & 76.8 & 90.6   & 94.2    & 98.5  & 84.4 & 0.0  & 37.7   & 59.3   & 74.0 & 83.1  & 91.6  & 77.4 & 77.2     & 78.8  & 60.6    \\
            \begin{tabular}[c]{@{}c@{}}StratifiedFormer  \cite{44qq}\end{tabular}                                                  & 72.0 & 78.1 & 91.5   & -       & -     & -   & -    & -      & -      & -    & -     & -     & -    & -        & -     & -       \\ \hline
            \begin{tabular}[c]{@{}c@{}}Point Transformer   (ours)\end{tabular}                                                        & 70.6 & 77.1 & 90.5   & 93.9    & 97.6  & 85.7 & 0.0  & 44.7   & 61.0   & \textbf{78.4} & 82.8  & 90.3  & 75.1 & 72.2     & 77.7  & 58.6    \\
            + PAGWN                                                                                                                     & 71.4 & 77.9 & 91.1   & 94.7    & 98.3  & 87.2 & 0.0  & 37.2   & 62.4   & 75.9 & 82.8  & 91.0  & 81.9 & 74.0     & \textbf{81.5}  & 60.7    \\
            \begin{tabular}[c]{@{}c@{}}StratifiedFormer   (ours)\end{tabular}                                                        & 71.7 & 77.6 & \textbf{91.9}   & \textbf{95.5}    & \textbf{98.7}  & \textbf{87.3} & 0.0  & 42.7   & 63.6   & 74.4 & 85.7  & 91.0  & 69.2 & 79.0     & 80.3  & \textbf{65.0}    \\
            + PAGWN                                                                                                                     & \textbf{72.2} & \textbf{78.2} & 91.4   & 95.0    & 98.1  & 85.9 & 0.0  & \textbf{48.7}   & 62.0   & 70.0 & 82.7  & \textbf{92.1}  & \textbf{84.4} & 78.3     & 76.8  & 64.0    \\
            \bottomrule
            \end{tabular}
            \label{tab:1}
        \end{table*}

        It can be seen that the results of our improved module are closer to the ground truth.
        From all three scenarios, we can find that the PAGWN module performs better in recognizing large contiguous regions such as the ceiling, wall, and window.
        In addition, our model has apparent advantages in recognizing small objects.
        With the proposed PAGWN module, Point Transformer can recognize the smoke detector on the ceiling, which could not be recognized before.
        Furthermore, only our proposed module can locate the clock, cabinet, and bookcase in all tested networks.
        We can also find that our proposed module has improved the recognition of object boundaries.
        The recognized objects have more precise boundaries and are closer to the ground truth, which can be seen from the frame in the second scene and the pillow in the third scene.
        All this shows that the PAGWN module can better balance and utilize local and global information.

        It can be found from Table \ref{tab:1} that the mIoU and mAcc of StratifiedFormer have increased while the OA has decreased.
        From each category's results, our module has significantly improved the recognition of sofa from 69.2\% to 84.4\% and column from 42.7\% to 48.7\%.
        However, the recognition of the board and door is not suitable.
        Considering that sofa and column are more complex categories to be recognized, the PAGWN module is better for hard-to-classify items than for easy-to-classify items.
        So, it leads to the rise of mAcc and the decline of OA.
        As shown in Figure \ref{fig:6} and Figure \ref{fig:10}, our method performs better in the boundary region where the $\sigma$ is larger, which is essential for point cloud understanding.

    \subsection{Semantic Segmentation on S3DIS (6-fold Cross-validation)}
        The 6-fold cross-validation experiments on S3DIS are conducted to prove our module's effectiveness further. 
        Point Transformer \cite{24Zhao_2021} and StratifiedFormer \cite{44qq} are also kept as the backbone. 
        We keep all settings and networks remained. Table \ref{tab:4} shows the details. 
        As can be seen, with the proposed PAGWN module, the performances have been raised to 77.6\%/85.8\%/91.7\% (mIoU/mAcc/OA). 
        It outperforms the best model PointNeXt-XL \cite{45https://doi.org/10.48550/arxiv.2206.04670} (74.9\%/83.0\%/90.3\% (mIoU/mAcc/OA)) by 2.7\% on mIoU and achieves state-of-the-art performance.
        \begin{table*}[!htb]\centering
            \renewcommand\arraystretch{1.1}
            \caption{Results of different models on S3DIS (6-fold cross-validation).}
        \begin{tabular}{c|ccc}
        \toprule
        Method & mIoU (\%) & mAcc (\%) & OA (\%) \\ \hline
        \begin{tabular}[c]{@{}c@{}}PointNet \cite{12Charles_2017}\end{tabular} & 47.6 & 66.2 & 78.5 \\
        \begin{tabular}[c]{@{}c@{}}PointWeb \cite{23Zhao_2019}\end{tabular}  & 66.7 & 76.2 & 87.3 \\
        \begin{tabular}[c]{@{}c@{}}KPConv \cite{31Thomas_2019}\end{tabular} & 70.6 & 79.1 & - \\
        \begin{tabular}[c]{@{}c@{}}Point Transformer \cite{24Zhao_2021}\end{tabular} & 73.5 & 81.9 & 90.2 \\
        \begin{tabular}[c]{@{}c@{}}PointNeXt-XL \cite{45https://doi.org/10.48550/arxiv.2206.04670}\end{tabular}  & 74.9 & 83.0 & 90.3 \\ \hline
        Point Transformer + PAGWN & 74.1 & 82.5 & 90.2 \\
        StratifiedFormer + PAGWN & \textbf{77.6} (+\textbf{2.7}) & \textbf{85.8} (+\textbf{2.8}) & \textbf{91.7} (+\textbf{1.4})\\
        \bottomrule
        \end{tabular}
        \label{tab:4}
        \end{table*}

\begin{figure*}[!htb]
            \centering
            \includegraphics[width=0.7\linewidth]{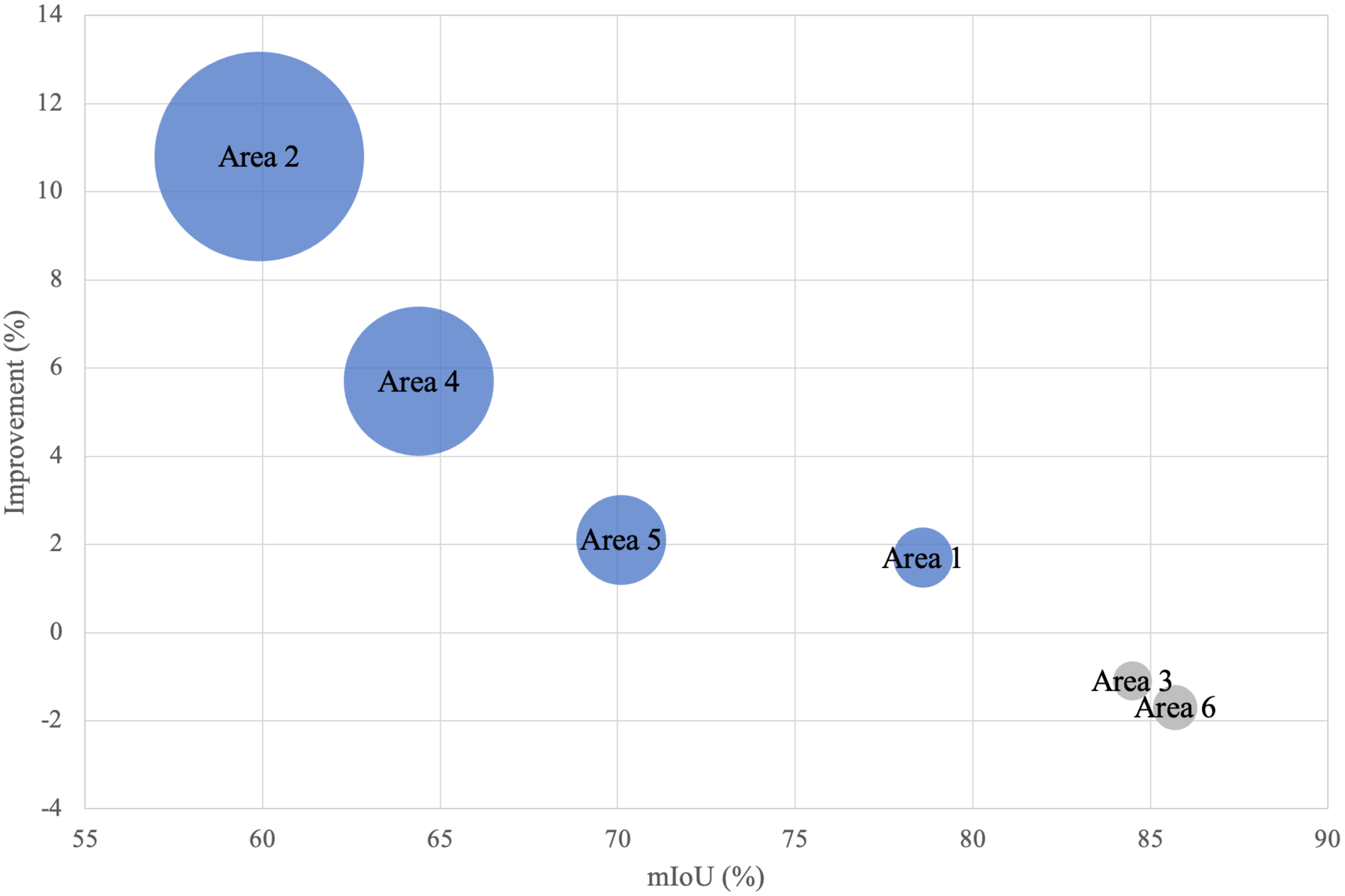}
            \caption{The effects we brought in different areas. There are six areas in the S3DIS dataset. Our method brings improvement to four of them whose accuracy is lower. As the difficulty of recognition increases, the improvement brought by our method also becomes larger. The mIoU of area 2 is originally 59.9\%, which is the lowest among all six areas, and our method brings an obvious improvement of 10.8\%. The blue ball means that we improve the accuracies and the gray one means the opposite. The bigger the ball is, the larger the effectiveness we bring.}
            \label{fig:11}
            \end{figure*}

        Figure \ref{fig:11} illustrates the effects brought in different areas. Table \ref{tab:6} shows the exact results. It can be find that among all six areas, our method brings improvement in four of them. These four areas are the four with lower accuracy. As the difficulty of recognition increases, the improvement brought by our method also becomes larger. The mIoU of area 2 is originally 59.9\%, which is the lowest among all six areas, and our method brings an obvious improvement of 10.8\%. This shows that our method is more effective for complex sample recognition, which is consistent with our previous conclusions.

        \begin{table*}[!htb]\footnotesize\centering
            \setlength{\tabcolsep}{2mm}
            \renewcommand\arraystretch{1.3}
            \caption{6-fold cross-validation results (mIoU, \%) on different areas in S3DIS.}
        \begin{tabular}{c|cccccc}
        \toprule
        Method & Area 1 & Area 2 & Area 3 & Area 4 & Area 5 & Area 6 \\ \hline
        \begin{tabular}[c]{@{}c@{}}PointNeXt-XL \cite{45https://doi.org/10.48550/arxiv.2206.04670}\end{tabular}  & 78.6 & 59.9 & \textbf{84.5} & 64.4 & 70.1 & \textbf{85.7} \\ \hline
        StratifiedFormer + PAGWN & \textbf{80.3} (+\textbf{1.7}) & \textbf{70.7} (+\textbf{10.8}) & 83.4 (-\textbf{1.1}) & \textbf{70.1} (+\textbf{5.7}) & \textbf{72.2} (+\textbf{2.1}) & \textbf{84.2} (-\textbf{1.5})\\
        \bottomrule
        \end{tabular}
        \label{tab:6}
        \end{table*}

    \subsection{Ablation Study}

        We use Point Transformer as the backbone to conduct ablation studies for each module. The necessity of each module is analyzed. Table \ref{tab:2} shows the results of different components.

        \begin{table*}[!htb]\centering

            \renewcommand\arraystretch{1.2}
        \caption{Ablation studies: Different components.}
        \begin{tabular}{l|ccc}
        \toprule
        Component                           & mIoU (\%)        & mAcc (\%)       & OA (\%)       \\ \hline
        Point Transformer                    & 70.4        & 76.5        & 90.8        \\
        + Window-Norm   and Pre-Abstraction & 70.9 (+0.5) & 77.1 (+0.6) & \textbf{91.1} (+0.3) \\
        + Group-wise                        & \textbf{71.4} (+1.0) & \textbf{77.9} (+1.4) & \textbf{91.1} (+0.3) \\
        \bottomrule
        \end{tabular}
        \label{tab:2}
        \end{table*}

        \textbf{Window-normalization and Pre-Abstraction}  The role of windownorm is to map the neighborhood into the spherical unit space. The neighborhood volume obtained by its $K$ nearest points differs for different center points. Therefore, the scale and density of the point cloud greatly influence the neighborhood. If the point cloud features are directly learned without normalization, different neighborhoods will probably have different volumes. Since neighborhoods with different sizes share the same network weights, the model sacrifices the accuracy in each region to keep the efficiency of the whole point cloud. However, windownorm can effectively solve this problem. Besides, it can also solve the problem of sampling offset in the sampling stage. The sampling region can be effectively constrained around the center point.

        Furthermore, it can be observed that the neighbor and center features have the same dimension. The Cartesian coordinates are used as a supplement to the spatial information of neighbor points, which makes the neighbor feature dimension larger than that of the center point. Neighbor information will have a greater weight in the network. In the downsampling stage, the network learns local features and maintains global features. The pre-abstraction module is used to pre-abstract the local feature, ensuring neighbor points have the same status as the center point. In this way, local and global features can achieve a better balance. With the constraints of the pre-abstraction module, windownorm can work better. The two modules cooperate to make the model perform better. As can be seen from Table \ref{tab:2}, the model accuracies are improved from 70.4\%/76.5\%/90.8\% (mIoU/mAcc/OA) to 70.9\%/77.1\%/91.1\% with windownorm and pre-abstraction modules. The improvement of model accuracy shows that both windownorm and pre-abstraction modules are effective.

        \textbf{Group-wise}  The points closer to the center point are considered to provide more local texture information. Meanwhile, points farther away from the center provide spatial information with a larger receptive field. We take the $m$ nearest points as the first set that provides local information. The remaining neighbor points serve as a second set that provides spatial information. Use group-wise windownorm for each of the two sets of points. A model that does not use a grouping strategy can be expressed as:
        \begin{equation}
          \begin{aligned}
      x_{i,j}^{*} = {L{B_2}\left( {\left[ {L{B_1}\left( {WN\left( {\left[ {{c_{i,j}},{x_{i,j}}} \right]} \right)} \right),{x_i}} \right]} \right)}\\
      {\hat x_{i}} = ReLU\left( {\mathop {{\rm{Maxpool}}}\limits_{j = 1, \cdots ,K} \left\{ x_{i,j}^{*} \right\}} \right); \forall {x_i} \in FPS\left( P \right); \\
      {\left( {{c_{i,j}},{x_{i,j}}} \right)} \in KNN\left( {{x_i}} \right),j = 1, \cdots ,K,
          \end{aligned}
      \end{equation}
        where WN means windownorm. As seen from Table \ref{tab:2}, the model performance has been further improved with the grouping strategy. The model accuracies are further improved from 70.9\%/77.1\%/91.1\% (mIoU/mAcc/OA) to 71.4\%/77.9\%/91.1\%.

        In addition, we also conduct experiments and analyses on the number of grouping points $m$. $m$ is used to determine the number of points involved in each type of information in the neighborhood. It should be noted that the number of grouping points $m$ may have different effects for different models and tasks. We do experiments to find optimal $m$ for our module. Table \ref{tab:3} shows the results of different $m$. It can be seen from Table \ref{tab:3} that the model accuracy is optimal with $m$=3.
        \begin{table}[!htb]\centering
            \renewcommand\arraystretch{1.1}
         \caption{Ablation studies: The number of grouping points $m$.}
        \begin{tabular}{c|ccc}
        \toprule
        $m$ & mIoU (\%) & mAcc (\%) & OA (\%) \\ \hline
        1 & 70.1 & 76.5 & 90.8 \\
        2 & 69.8 & 76.3 & 90.8 \\
        3 & \textbf{71.4} & \textbf{77.9} & \textbf{91.1} \\
        4 & 70.2 & 77.0 & 90.7 \\
        \bottomrule
        \end{tabular}
        \label{tab:3}
        \end{table}

    \subsection{Analysis of Sampling Complexity}

        The PAGWN module has one Linear layer to aggregate location information and neighbor features. Benefiting from parallel operations, the normalization process does not affect computational efficiency. In our proposed method, only a few learnable parameters are added. Compared to the original method, the PAGWN module has a small increment in memory usage and inference time. Table \ref{tab:5} compares complexity between different models on S3DIS.
        \begin{table}[!htb]\centering
          \setlength{\tabcolsep}{3mm}
            \renewcommand\arraystretch{1.1}
         \caption{Complexity comparison on S3DIS (Area 5).}
        \begin{tabular}{c|cc}
        \toprule
        Method                           & Point Transformer & PAGWN     \\ \hline
        Parameters       & 7,767,729        & 8,031,729 \\
        Video memory & 10175m           & 11897m    \\
        Training time  & 34min10s         & 35min47s  \\
        Inference time      & 30min24s         & 31min45s  \\
        \bottomrule
        \end{tabular}
        \label{tab:5}
        \end{table}

\section{Conclusions}
    \label{sec:4}
    In this work, we try to improve the information acquisition rate in the downsampling stage.
    In particular, we propose the group-wise window-normalization module to unify inconsistent point densities and obtain multi-type information, including texture and spatial information.
    Furthermore, the pre-abstraction strategy is leveraged to balance local and global features.
    With the proposed PAGWN module, the recognition of the sofa and column is improved from 69.2\% to 84.4\% and from 42.7\% to 48.7\%, respectively.
    The benchmarks of segmentation on S3DIS (Area 5) are improved from 71.7\%/77.6\%/91.9\% (mIoU/mAcc/OA) to 72.2\%/78.2\%/91.4\%.
    The accuracies of 6-fold cross-validation on S3DIS are 77.6\%/85.8\%/91.7\%.
    It outperforms the best model PointNeXt-XL \cite{45https://doi.org/10.48550/arxiv.2206.04670} (74.9\%/83.0\%/90.3\%) by 2.7\% on mIoU and achieves state-of-the-art performance.
    Extensive experiments show that the PAGWN module has the following advantages:
    1) It is computationally efficient with a small number of learnable parameters.
    2) The normalization of inconsistent point densities is essential to point cloud understanding.
    3) Texture and spatial information are extracted to enhance the network performance.
    4) Local and global information can be well balanced.
    5) The proposed module performs better on small object recognition, and the results have more precise boundaries than others.

{\small
\bibliographystyle{ieee_fullname}
\bibliography{references}
}

\end{document}